\documentclass{article}
\usepackage{jmlr2e-arxiv}

\usepackage{algorithm}
\usepackage{algorithmic}
\usepackage{bbm}
\usepackage{graphicx} 
\usepackage{amsmath}
\usepackage{amssymb}
\usepackage{color}
\usepackage{verbatim}

\renewcommand{\comment}[1]{}

\newcommand{\1}{\mathbf{1}}

\newcommand{\beq}{\begin{equation}}
\newcommand{\eeq}{\end{equation}}
\newcommand{\beqa}{\begin{eqnarray}}
\newcommand{\eeqa}{\end{eqnarray}}

\newcommand{\rf}{\mathbf{r}}
\newcommand{\Ho}{\mathcal{H}}
\newcommand{\RR}{\mathbb{R}} 
\newcommand{\M}{\mathcal{M}}
\newcommand{\wcut}{\ensuremath{W\!Cut}}
\newcommand{\qed}{\nobreak \ifvmode \relax \else
      \ifdim\lastskip<1.5em \hskip-\lastskip
      \hskip1.5em plus0em minus0.5em \fi \nobreak
      \vrule height0.75em width0.5em depth0.25em\fi}

\newlength{\picwi}
\begin{document}

\title{Estimating Vector Fields on Manifolds and the Embedding of Directed Graphs}

\author{Dominique Perrault-Joncas%
\thanks{DPJ is with Amazon.Inc, Seattle, USA.}, 
       Marina Meil\u{a}%
\thanks{MM is with the Department of Statistics, University of Washington, Seattle, USA.}
}

\maketitle

\begin{abstract}
This paper considers the problem of embedding directed graphs in
Euclidean space while retaining directional information. We model a
directed graph as a finite set of observations from a diffusion on a
manifold endowed with a vector field. This is the first generative
model of its kind for directed graphs. We introduce a graph embedding
algorithm that estimates all three features of this model: the
low-dimensional embedding of the manifold, the data density and the
vector field. In the process, we also obtain new theoretical results
on the limits of ``Laplacian type'' matrices derived from directed
graphs. The application of our method to both artificially constructed
and real data highlights its strengths.
\end{abstract}

\section{Motivation}
\label{sec:intro}
Recent advances in graph embedding and visualization have brought to
the foreground the graph Laplacian, as a key element to the analysis.
Its properties, and in particular it's convergence as the sample size
goes to infinity, put the study of graph embedding on a strong
statistical foundation, and lead to particularly elegant
algorithms. This direction of research has led to important
developments in manifold learning~\cite{belniy02} and diffusion
maps~\cite{fp05}.

However, the graph Laplacian and related algorithms are defined for
undirected graphs. There are many instances of graph data, such as social
networks, alignment scores between biological sequences, international
relations, and citation data,  which are naturally
asymmetric. A simple approach for embedding or clustering this type of
data is to disregard the asymmetry by studying the spectral properties
of $A+A^T$ or $A^TA$, where $A$ is the affinity matrix of the
graph. This can work for data with weak assymetry, or when we are only
interested in a limited analysis of the graph's properties. 

Some authors have proposed methods to preserve the asymmetry
information contained in data: \cite{penmei07} is an interesting
generalization of the popular (multiway) Normalized Cut that captures
part of the asymmetry of the data. The works of \cite{zhou:05a},
\cite{zhou:05b}, \cite{chung:conf} propose principled way to obtain a
symmetric similarity from asymmetric graph data. In all these works,
the resulting symmetric matrix can be used for both embedding and
clustering.  Although quite successful, these works adopt a purely
graph-theoretical point of view. Thus, they are not concerned with the
generative process that produces the graph, nor with the statistical
properties of their algorithms.

In the context of social networks~\cite{hoffhanraf02} proposes latent
space model with actor-specific ``activity'' that can account
differences in affinity between two actors which are otherwise
symmetric. There is an assumed statistical model for the
graph, but of a finite type, the model for the asymmetry {\em is rather
ad-hoc. }

The present paper takes steps beyond these existing works, by
considering a generative process for directed graphs that explicitly
contains geometry represented by a manifold in Euclidean space, a
sampling distribution on this manifold, and a vector field that
accounts for the observed directionality of the pairwise relations.

We view the nodes of the given directed graph as a finite sample from
this manifold, and the (weighted) links as macroscopic observations of
an infinitesimal diffusion process on the manifold. We derive how this
continous process determines the overall connectivity and asymmetry of
the resulting graph. Important insights come from finding the
continous limits of Laplace-type operators on digraphs. These will be
presented in the following two sections.

Based on these asymptotic results, in Section \ref{sec:recovery}, we
derive algorithms that, in the limit, recover the generative process:
manifold geometry, data density, and local directionality, up to their
intrinsic indeterminacies. We also pay attention, in Section
\ref{sec:wcut}, to what the limit process tells us about the attributes
of related algorithms such as the MNCut and more particularly its
extension to directed graphs, the WCut \cite{penmei07}. This also
helps situate our algorithm in the context of previous work.

Section \ref{sec:experiments-artif} validates our methods on
artificial data, while Section \ref{sec:experiments-real} analyzes in
depth a real example. Discussion of the relation with previous work is
presented in Section \ref{sec:discussion}.

\section{Model and Problem Definition} \label{sec:mod_prob}

We assume that we are given a directed graph $G$ with $n$ nodes, and
with edge {\em weights} (or {\em affinities}) given by the $n\times n$
matrix $A$, with $A_{i,j}\geq 0$ denoting the affinity between node
$i$ and node $j$.

According to our generative paradigm, we assume that the graph $G$ is
sampled from an unobserved compact, closed and smooth manifold $\M$ of
dimension $d$. The sampling need not be uniform; we assume that it is
done according to some distribution $p$ on $\M$, non-zero
everywhere. Under this last assumption, we find it convenient to
represent $p$ by a {\em potential function} $U$, with
$p(x)=\exp(-U(x))$ on $\M$. These assumptions are similar to those
made for modeling undirected graphs
in~\cite{dmap06,dsys06,fp05}. Because $G$ is a directed graph, the
edge weights $A_{i,j}$ and $A_{j,i}$ between nodes $i$ and $j$ are
assigned by an asymmetric similarity kernel $k_{\epsilon}(x,y)$, with
{\em bandwidth} $\epsilon$ which will be defined in more detail
shortly. Further, we assume that the asymmetry of $k_{\epsilon}(x,y)$
originates from a vector field $\mathbf{r}(x)$ on $\mathcal{M}$, which
assigns a prefered direction between weights $a_{i,j}$ and
$a_{j,i}$. The schematic of this sampling process is shown in the top
left corner of Figure ~\ref{fig:mod_sem} below.

\begin{figure}
\begin{center}
\centerline{\includegraphics[width=0.6\columnwidth]{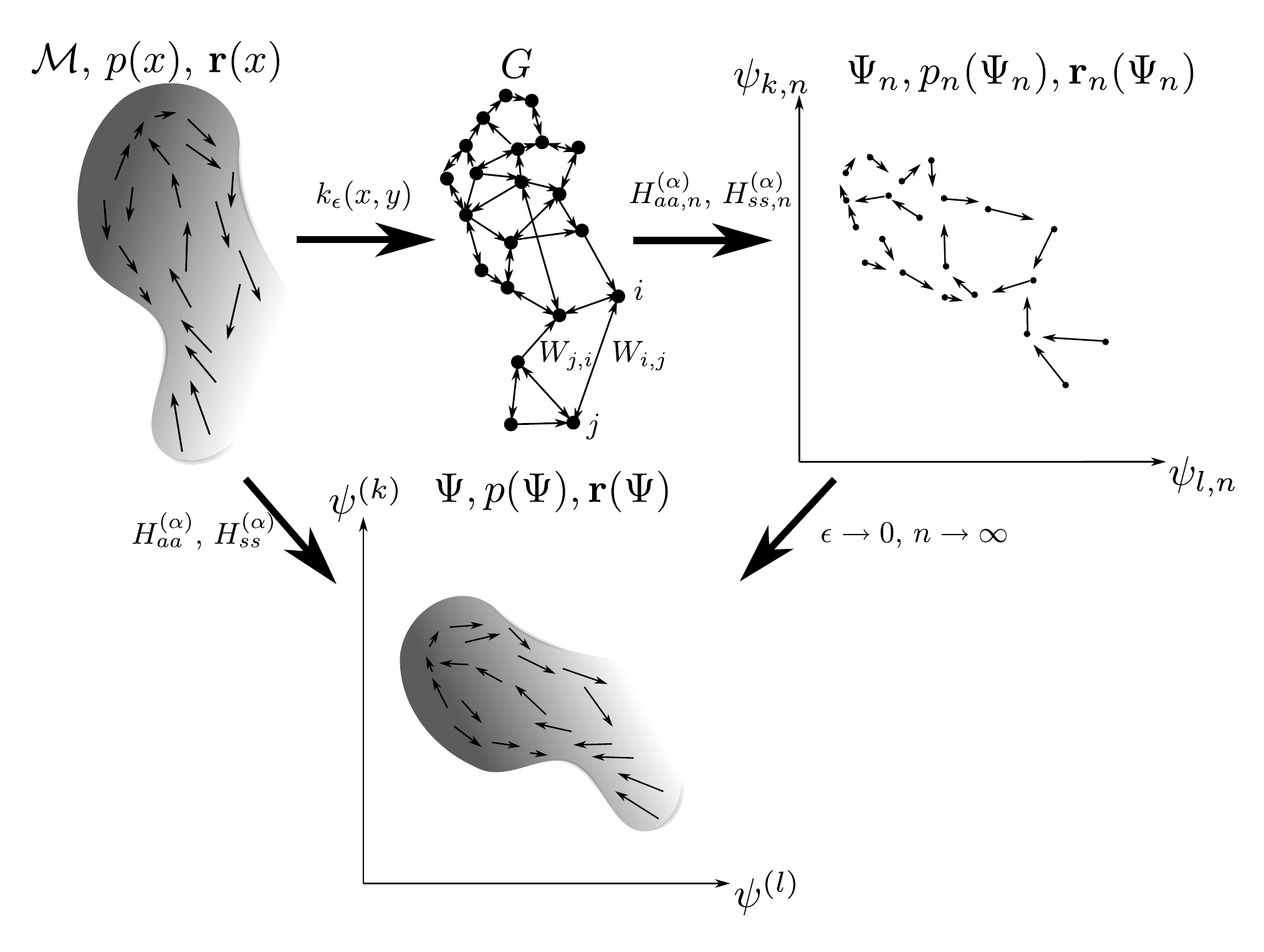}}
\caption{From left to right, the schematic shows the graph generative process mapping $\mathcal{M}$, $e^{-U(x)}$, and $\mathbf{r}(x)$ to $G$ and the subsequent embedding $\Psi_n$ of $G$ by the eigenvectors of $H_{n,ss}^{(1)}$ and operators $H_{n,aa}^{(\alpha)}$, $H_{n,ss}^{(\alpha)}$ defined in section~\ref{sec:H_ops} below. As $\epsilon\rightarrow 0$, and $n\rightarrow \infty$ the embedding converges and preserves the geometry of $\mathcal{M}$, the distribution $p=e^{-U(x)}$, and the directionality $\rf(x)$ }
\label{fig:mod_sem}
\end{center}
\vskip -0.2in
\end{figure} 

The question is then if the generative process described above can be
recovered in all its components, $\M$, distribution $p=e^{-U(x)}$, and
directionality $\mathbf{r}(x)$, be recovered from $G$? Moreover, is
there a practical algorithm to do so?  And is this process also
consistent as sample size increases from $n\rightarrow
\infty$? 

In the case of undirected graphs, the theory of Laplacian
eigenmaps~\cite{belniy02} and Diffusion maps~\cite{dmap06} answers
this question in the affirmative\footnote{The embedding obtained is up
to a smooth invertible transformation, but is not an {\em
isometry}}. That is, the local features of the geometry of
$\mathcal{M}$ and $p=e^{-U(x)}$ can be inferred using spectral graph
theory. The aim here is to build on the undirected problem and recover
all three components of the generative process from a directed $G$.

\comment{
The spectral approach to undirected graph embedding relies on the fact
that eigenfunctions of the Laplace-Beltrami are known to preserve the
local geometry of $\mathcal{M}$~\cite{belniy02}. With a consistent
empirical Laplace-Beltrami operator based on $G$, its eigenvectors
also recover the local geometry of $\mathcal{M}$ and the eigenvectors
converge to the corresponding eigenfunctions on $\mathcal{M}$ as
$n\rightarrow \infty$. For a directed graph $G$, things are slightly
more complicated as recovering directionality requires a more complete
knowledge of the geometry than what Laplacian eigenmaps can
provide. Nevertheless, the underlying principle remains the same in
that we can define operators that recover the local directional
component $\mathbf{r}(x)$ provided the geometry is known up to an
isometric scaling.}\comment{ The schematic for this is shown in
Figure~\ref{fig:mod_sem} where Multidimensional Scaling (MDS) and
operators, $\mathcal{H}_{n,1}^{(\alpha)}$ and
$\mathcal{H}_{n,4}^{(\alpha)}$, are used to obtain the embedding
$\Phi_n$, distribution $p_n$, and vector field $\mathbf{r}_n(x)$. As
the sample size increases, this embedding converges to $\Phi$, $P$,
and $\mathbf{r}(x)$, where $\Phi$ is an isometric scaling of the
original manifold $\mathcal{M}$. 
of $\mathcal{H}_{n,1}^{(\alpha)}$ and $\mathcal{H}_{n,4}^{(\alpha)}$,
$\Phi_n$, $P_n$, and $\mathbf{r}_n(x)$ converge to $\Phi$, $P$, and
$\mathbf{r}(x)$, where $\Phi$ is the local geometry preserving
embedding of $\mathcal{M}$ into $\RR^m$. }

Now we proceed to define our framework in more detail.

\subsection{Choice of Kernel} \label{sec:kernel}

The starting assumption in our model is that the observed affinities
between nodes $x,y$ in $G$ are given by the (asymmetric) {\em
diffusion kernel} $k_\epsilon(x,y)$, a smooth function on $\M\times
\M$. We start by decomposing the asymmetric affinity kernel
$k_{\epsilon}(x,y)$ into its symmetric
$h_{\epsilon}(x,y)=h_{\epsilon}(y,x)$ and anti-symmetric
$a_{\epsilon}(x,y)=-a_{\epsilon}(y,x)$ parts. This decomposition is
unique. 
\begin{equation}
k_{\epsilon}(x,y)=h_{\epsilon}(x,y)+a_{\epsilon}(x,y)\, .\label{eq:kform}
\end{equation} 
The symmetric component $h_{\epsilon}(x,y)$ is assumed to satisfy the
following properties: 
\begin{enumerate} \setlength{\parskip}{0em}
\item $h_{}(x,y) = h_{}(||y-x||^{2})$.
\item $h_{}(u)\ge0$ for $u\geq 0$. 
\item $h_{\epsilon}(x,y)=\frac{h_{}(||y-x||^{2}/\epsilon)}{\epsilon^{d/2}}$ \label{it:eps}
\item (Simplifying assumption: $h_{}(u)\propto e^{-u}$)  \label{it:exp}
\end{enumerate}
As a reminder, $d$ in assumption \ref{it:eps} above is the dimension
of the manifold. This form of symmetric kernel was used
in~\cite{dmap06} to analyze the diffusion map.  Assumption
\ref{it:exp} was only required for large values of $u$ in that
paper. We use it to simplify the presentation in Sections
\ref{sec:asymptotics} and \ref{sec:recovery} but it is not essential.

To the symmetric kernel $h_\epsilon$, we add an {\em asymmetric}
component $a_{\epsilon}(x,y)=-a_{\epsilon}(x,y)$ chosen so that
$k_{\epsilon}(x,y)\ge0$. This decomposition has the advantage that it
has a direct correspondence to the graph adjacency matrix
decomposition into symmetric and skew-symmetric parts.
\begin{equation}\label{eq:akern}
a_{\epsilon}(x,y)
\,=\,\frac{\mathbf{r}(x,y)}{2}\cdot(y-x)h_\epsilon(x,y)
\end{equation}
\comment{Equation~\eqref{eq:kform} is a good starting point in defining a family of kernels that can generate any directed graphs and are easy to interpret. However, it is convenient to consider something a little more specific in order to facilitate the asymptotic derivation. We retain the radial symmetry of of $h_{\epsilon}$ and assume}
with $\rf(x,y) = \rf(y,x)$ so that $a_{\epsilon}(x,y)=-a_{\epsilon}(y,x)$
is enforced. Here $\rf(x,y)$ is a smooth vector field on the manifold
that gives an orientation to the asymmetry of the kernel
$k_{\epsilon}(x,y)$. It is worth noting that the dependence of
$\rf(x,y)$ on both $x$ and $y$ implies that the domain of $\rf (x,y)$ is
$\M\times\M$; however, the dependence in $y$ is only important {\em
locally}, as will be made evident in Section \ref{sec:asymptotics} below, and as such it is appropriate to talk about $\rf (x,y)$ being a vector field
on $\M$.

It is worth pointing out that even though the form of
$a_{\epsilon}(x,y)$ might seem restrictive at first, it is
sufficiently rich to describe any smooth vector field $\mathbf{w}(x)$. This can
be seen by taking $\rf (x,y) = \mathbf{w}(x)+
\mathbf{w}(y)$ so that at $x=y$ the vector field is given by $\rf
(x,x) = 2\mathbf{w}(x)$. In addition, as it will become clear in
Section \ref{sec:asymptotics}, $\rf (x,y)$ will only be identifiable at $x=y$,
i.e. effectively as the vector field $\rf (x,x)$. 

The above form of $a_{\epsilon}(x,y)$ also means that (proof
omitted)\begin{equation}
\lim_{\epsilon\to 0}\int_{\M}a_{\epsilon}(x,y)f(y)dy=0\;\text{for}\;f\in {\cal C}^2(\M).\end{equation}
This property will be exploited in  our main Theorem \ref{thm:lim}.

\subsection{From kernel to transport operator}

A diffusion kernel $k_\epsilon(x,y)$ represents the transport between
$x$ and $y$ in the time interval $\epsilon$ \cite{dmap06}. Starting from it,
one can define the {\em density after time $\epsilon$} as 
\begin{equation} \label{eq:pop}  
p_{\epsilon}(x)=\int_{\mathcal{M}}k_{\epsilon}(x,y)p(y)dy\,.
\eeq
Furthermore, one defines a natural {\em transport operator} on $\M$ based on
$k_{\epsilon}(x,y)$ as
\beq \label{eq:kernop}
T_{\epsilon}[\phi](x)=\int_{\mathcal{M}}\frac{k_{\epsilon}(x,y)p(y)}{p_{\epsilon}(x)}\phi(y)dy\, 
\end{equation}
where $k_\epsilon(x,y)p(y)/p_{\epsilon}$ is the {\em transition
  density} after time $\epsilon$ and $T_{\epsilon}[\phi](x)$
represents the diffusion of a distribution $\phi(y)$ through the
transition density.  The attentive reader will have recognized in this
infinitesimal operator the continous limit of the transition
probability matrix $P=D^{-1}A$ given by normalizing the adjency matrix
$A$ of $G$ by $D=\text{diag}(A \1)$ (where $\1$ denotes the vector
with all elements equal to one). Correspondingly, the eigenfunctions
of $T_\epsilon$ are limits for the eigenvectors of $P$ as shown by
\cite{belniy02}.

The diffusion process over $\mathcal{M}$ based on kernel
$k_{\epsilon}$ is defined by its (backward\footnote{A forward
  transport operator and equation can be derived for the same process,
  but  are not presented here as the backward equation will be
  sufficient for our purposes.}) equation
\begin{equation}
\frac{\partial\phi}{\partial t}=\lim_{\epsilon\to0}\frac{(T_{\epsilon}-I)\phi}{\epsilon}\,.\label{eq:infop}
\end{equation}

With these definitions, the strategy we will follow is first, to study
the limits of transport operators like the ones defined above when
$\epsilon\, \longrightarrow\,0$. These limits will be other operators
on $\M$, which for convenience we call ``Laplacian type'' operators.
If the diffusion kernel $h$ has been chosen judiciously, the limit
operators will not depend on $h$, nor on $\epsilon$ but will depend on
the other features of the generative process, i.e. the manifold $\M$,
the sampling density $p$ and the vector field $\rf$. We will study if
any combination of these operators has eigenvectors that can separate
out one of these components. This has already been done by
\cite{dmap06} to separate $\M$ and $p$, in the absence of $\rf$. If
such operators are found, we will apply their discrete counterparts to
the graph $G$ to obtain the estimates of $\M$, $p$ and $\rf$ from the
observed data. The next section will derive the core result regarding
the limits when $\epsilon\rightarrow 0$. The following sections
will show how this result can be exploited to obtain the desired
spectral estimation algorithm, as well as to analyze a number of
related spectral methods for directed graphs.

\section{Continuous Limit of Laplacian Type Operators}
\label{sec:asymptotics}

We now derive the main asymptotic result from which all the
infinitesimal transitions of interest can be obtained.
This derivation and its application to obtaining closed form
expressions for various operators on $\M$ are the subject of this
section.

\begin{theorem} \label{thm:lim}
Let $\M$ be a compact, closed smooth manifold and $k_{\epsilon}(x,y)$
an asymmetric similarity kernel statisfying the conditions of
section~\ref{sec:mod_prob}. Then for any function $f\in C^2(\M)$, the
integral operator based on $k_{\epsilon}$ as the asymptotic expansion
\begin{equation}
\int_{\M}k_{\epsilon}(x,y)f(y)dy=m_{0}f(x)+\epsilon g(f,x)+o(\epsilon)\,,\label{eq:asaim}
\end{equation}
where
\beqa
g(f,x)&=&\frac{m_{2}}{2} \left( \omega(x)f(x)+\Delta f(x)+\rf \cdot\nabla f(x)\right.
\label{eq:w} \\
&&\left. + f(x)\nabla \cdot \rf +c(x)f(x) \right)
\,, \nonumber
\eeqa
and
\begin{equation}
m_{0}=\int_{\RR ^{d}}h(||u||^{2})du\,, \quad
m_{2}=\int_{\RR^{d}}u_{i}^{2}h(||u||^{2})du\,. \nonumber
\end{equation}
\end{theorem}
In the above, $u_i$ is any one of the $d$ coordinates of $\RR^d$. 

From Theorem \ref{thm:lim} it follows immediately that
\begin{corollary}
Under the conditions of Theorem \ref{thm:lim}, and assuming $m_0=1$ and $m_2=2$, we have that
\beq 
\lim_{\epsilon\longrightarrow 0}\frac{T_\epsilon-I}{\epsilon}[\phi]
\;=\;g(\phi,\cdot)\;\equiv\;\omega\phi+\Delta\phi+\rf\cdot \nabla\phi+\phi\nabla\cdot\rf+c
\eeq
\end{corollary}
The proof of this expansion can be found in the Appendix. Here we will
only consider the origin and interpretation of the terms
in~\eqref{eq:w}, to help with the interpretation of some of the coming
results.

The origin of $\rf$, the vector field on $\M$, is quite
obvious as it is part of the explicit definition of $k_{\epsilon}$
(note that in the limit $\rf$ appears as depending on $x$ only), while
$\Delta f(x)$ is just the diffusion of $f(x)$ around point $x$. 

There remain to clarify the unknown functions $c(x)$ and
$\omega(x)$. Both of these are associated with the curvature of
$\M$. Specifically, $\omega(x)$ corresponds to the interaction between
kernel and $\M$ while, $c(x)$ corresponds to the interaction between
$\M$ and the component of $\rf$ perpendicular on $T_x\M$, the tangent
space of $\M$ at $x$.  Because they are interaction terms depending on
several unknown model components, we expect them to be difficult to
estimate, or even non-identifyable. Therefore, in what follows, we
will consider composite operators derived from the above, where these
terms cancel.

We mention that Assumption \ref{it:exp} is not essential for this
proof, but it is convenient for the simple form of the constants
$m_0=1$ and $m_2=2$ that one obtains under it.  Weaker assumptions,
similar to those in \cite{coif05} or \cite{ting10} are used in the
proof.

\subsection{Renormalized Limit Operators} \label{sec:H_ops}

Using on Theorem \ref{thm:lim}, it is now possible to derive the limit
forms of a number of operators on $\M$, the most interesting
being the renormalized diffusion operators introduced by~\cite{dmap06}
\begin{equation}\label{eq:ke}
k_{\epsilon}^{(\alpha)}(x,y)=\frac{{k_{\epsilon}(x,y)}}{p_{\epsilon}^{\alpha}(x)p_{\epsilon}^{\alpha}(y)} \;,\text{with}\,\alpha \in [0,1]. 
\end{equation}

We briefly present the key ingredients of these derivations.
Renormalizing $k_\epsilon$ as in \eqref{eq:ke}, means that we first
need the ``outdegrees'' of $k_{\epsilon}$ as
$p_{\epsilon}(x)=\int_{\M}k_{\epsilon}(x,y)p(y)dy$.  From Theorem
\ref{thm:lim}, taking $m_{0}=1$ and $m_{2}=2$ for simplicity, we
have: 
\begin{eqnarray}
 p_{\epsilon}\!\!\!& =\!\!\! & p+\epsilon(\omega p+\Delta p+2\rf\cdot\nabla p+2p\nabla\cdot \rf +c p ) +o(\epsilon)\,, \nonumber \label{eq:pasym}\\
\!\!\!&=\!\!\!&p[1+\epsilon(\omega+\frac{\Delta p}{p}+2\rf\cdot\frac{\nabla p}{p}
 +2\nabla\cdot \rf+c)]+o(\epsilon)\nonumber\\
\frac{1}{p_{\epsilon}} \!\!\!& = &\!\!\! \frac{1}{p}(1-\epsilon(\omega+\frac{\Delta p}{p}+2\rf\cdot\frac{\nabla p}{p}
 +2\nabla\cdot \rf+c))+o(\epsilon)\,. \label{eq:pe}
\end{eqnarray}
With the help of equations~\eqref{eq:pe} and~\eqref{eq:ke}, four
families of diffusion processes of the form
$\phi_t=\Ho_{m}^{(\alpha)}[\phi](x)$ can be derived, based on whether
$p_\epsilon$ are the outdegrees of $k_\epsilon$ or $h_\epsilon$ and
wether we renormalize $k_{\epsilon}$ or $h_\epsilon$ in~\eqref{eq:ke}.
The derivations involve applying Theorem \ref{thm:lim}, differential
calculus, and dropping all the terms containing $\epsilon$.

Specifically, if we use $k_{\epsilon}$ and its outdegrees $p_{\epsilon}$, we get the advected diffision equation: 
\begin{equation}
\Ho_{aa}^{(\alpha)}[\phi]=\Delta\phi-2\left(1-\alpha\right)\nabla\phi\cdot\nabla U-2 r\cdot(\nabla\phi) \,.\label{eq:r_alone}
\end{equation}
In general, this operator is not hermitian and so it commonly has complex eigenvectors. Nevertheless, $\Ho_{aa}^{(1)}$ will play an important role in extracting the directionality of the data.  

Meanwhile, if  $p_{\epsilon}$ are the outdegrees of $k_{\epsilon}$ but we renormalize 
$h_{\epsilon}$ instead, we recover a generalized version of the WCut diffusion of \cite{penmei07}\begin{equation}
\Ho_{as}^{(\alpha)}[\phi]=\Delta\phi-2\left(1-\alpha\right)\nabla\phi\cdot\nabla U -\phi(c+2\nabla\cdot r+2(\alpha-1)r\cdot\nabla U) \,.  \label{eq:WCut}
\end{equation}
This operator will be further discussed in Section \ref{sec:wcut}.

If we use the symmetric part $h_{\epsilon}$ to define the outdegrees $p_{\epsilon}$ but renormalize $k_{\epsilon}$ we obtain:
\begin{eqnarray}
\Ho_{sa}^{(\alpha)}[\phi]&=&\Delta\phi-2\left(1-\alpha\right)\nabla\phi\cdot\nabla U+2r\cdot\nabla\phi \nonumber\\
&& +c\phi-2(1-\alpha)\phi(r\cdot\nabla U)+2\phi(\nabla\cdot r) \,.  \nonumber \label{eq:thirdnorm}
\end{eqnarray}
Since $H_{sa}^{(\alpha)}$ is merely a combination of $H_{aa}^{(\alpha)}$
and $H_{as}^{(\alpha)}$, it adds little to the analysis. 

Finally, if we only consider the symmetric components for
$h_{\epsilon}$ and its outdegrees $p_{\epsilon}$, we recover the theory for symmetric graphs from \cite{dmap06}.
\begin{equation}
\Ho_{ss}^{(\alpha)}[\phi]=\Delta\phi-2(1-\alpha)\nabla\phi\cdot\nabla U.\label{eq:grad_alone-1}
\end{equation}
 Specifically when $\alpha=1$, the above becomes the Laplace-Beltrami operator 
\beq \label{eq:h4^1}
\Ho_{ss}^{(1)}[\phi]\;=\;\Delta\phi\,,
\eeq
and for $\alpha=0$, $\Ho_{ss}^{(0)}[\phi]$ is the limit of the MNCut
operator as shown by \cite{coif05}. The same authors showed that
\eqref{eq:h4^1} can be used to  separate the manifold from the
probability distribution~\cite{coif05}, a procedure we will use as
part of the algorithm presented in Section \ref{sec:recovery}.

\subsection{Continous limit of the WCut Operator}\label{sec:wcut}

In \cite{penmei07} is introduced the {\em Weighted Cut} \wcut~ a
directed version of the well known (multi way) Normalized Cut
criterion for clustering directed graphs. As it is well known
\cite{ShiMalik_ncut_pami:00,MShi:aistats01} clustering by the
Normalized Cut uses the eigenvectors of the {\em normalized graph
  Laplacian}; similarly the \wcut~ criterion is approximately
minimized by the eigenvectors of a symmetric matrix, which plays the
role of {\em directed graph Laplacian}. More precisely, the
``Laplacian'' $W$ used by the \wcut~ algorithm is given by
\beq \label{eq:wcutlt}
W=\frac{1}{2}(B+B^T),\;\text{with}\,B\,=\,T^{-1/2}(D-A)T^{-1/2} \eeq
where $A$ represents the affinity matrix, $D$ is the diagonal matrix
of out-degrees (row sums of $A$), and $T$ is a diagonal matrix of
positive node weights, set by the user. The weights in $T$ give the
algorithm's name. When the graph is dense enough that no out-degree is
0 or too small, a natural choice is to set $T_i$ equal to the
out-degree of node $i$, i.e. $T=D$. In this case, \eqref{eq:wcutlt}
simplifies to \beq \label{eq:wcutl}
W=I-D^{-1/2}\frac{A+A^T}{2}D^{-1/2}\,.  \eeq This form corresponds to
the normalized graph Laplacian when $A$ is a symmetric matrix. We will
now derive the limit of the Laplacian $W$ defined by \eqref{eq:wcutl}.

It is immediate from \eqref{eq:wcutl}, that the symmetric part of the kernel $h_{\epsilon}(x,y)$ is normalized by the outdegrees of the full kernel~\eqref{eq:pasym}. 
\beq
 \Ho_{as}[\phi](x) \, = \, \int_{\M}\frac{h_{\epsilon}(x,y)}{p_{\epsilon}(x)}\phi(y)p(y)dy\,.
\eeq
Using \eqref{eq:pe} and the fact that $\nabla p/p=-\nabla U$ and $(\Delta p)/p=-\Delta U-||\nabla U||^2$, after some cancellations we obtain \eqref{eq:WCut}. If we take $\alpha=0$ (kernel not renormalized) then the infinitezimal operator of \wcut~ corresponds to the PDE
\beq
\frac{\partial\phi}{\partial t}=\Delta\phi-2\nabla\phi\cdot\nabla U-(c-2 \rf\cdot\nabla U+2\nabla\cdot \rf)\phi\,. \label{eq:pdewcut}
\end{equation}
This equation describes an advected diffusion process, where the last
term on the right side acts as source/sink, depending on its sign. It
is worth pointing out that this term is a contribution of the
asymmetry of the kernel in the absence of which~\eqref{eq:pdewcut}
would be the diffusion equation for the MNCut operator.

Let us further simplify the PDE by assuming now that the field is
tangential, which cancels $c$, and setting $\nabla U=0$ (i.e. uniform
density), which cancels the diffusion effect caused by
the non-uniform density. Then, the source/sink term is
$-(\nabla\cdot\rf)\phi$. Remembering now that large negative values in
the divergence of a vector field mark the regions where the flow
accumulates or compresses, we recognize that this term will account
for the clustering effect due to the $\rf$ the directional flow on
$\M$. Thus, we have obtained an alternative motivation for the \wcut~
algorithm. It is one of the few times when one can theoretically
separate, under the manifold hypothesis, whether a spectral algorithm
is ``good'' for clustering or for embedding. In the present case, we
have shown that \wcut~ will generate distortions if used to obtain a
smooth embedding of a directed graph (unlike the algorithm in the next
section), but the effects of these distortions will be benefic for
clustering the graph nodes.

\section{The recovery algorithm}
\label{sec:recovery}

We are trying to recover three things from the generative process
here, the geometry/manifold $\mathcal{M}$, the density distribution
$p(x)=\exp(-U(x))$, and the vector field $\rf(x)$. 

The geometry of the data can be recovered from
\eqref{eq:grad_alone-1}, but only {\em locally}. As previously noted,
$\Ho_{ss}^{(\alpha)}$ with $\alpha = 1$ is the Laplace-Beltrami
operator whose eigenfunction recover the geometry of
$\mathcal{M}$~\cite{coif05}. Similarly, $p(x)$ can be obtained from
the forward adjoint operator of $\Ho_{ss}^{(1)}$ as the stationary
distribution of the resulting diffusion process. These operations are
described in steps 1.--6. (coordinates) and 7. (sampling density) of
Algorithm \ref{alg:geo}. In the algorithm, the matrices $Q,\,V$ and
$Q^{(1)}$ are respectively the discretized versions of
$p_\epsilon,\,k^{(1)}_\epsilon$ and the outdegrees of 
$k^{(1)}_\epsilon$. The $H_{aa}^{(1)}$ matrix is the discrete version of $\Ho_{ss,n}^{(\alpha=1)}$ from Figure \ref{fig:mod_sem}.

Step 6 takes the $d+1$ principal eigenvectors of $H_{aa}^{(1)}$. This
being a stochastic matrix, the first eigenvalue is 1, and its
eigenvector is the constant 1 vector\footnote{Assuming the graph is
connected.}. These are discarded, and the remaining $d$ eigenvectors
form the embedding coordinates, as shown in \cite{coif05} and
\cite{MShi:aistats01}, i.e. with $\Phi_{i,j}$ being the $j$-th coordinate of point $i$ in the data.

There is now only one feature of the graph that we are left to
recover, the vector field $\rf(x)$.  For this we isolate the term
$\rf\cdot\nabla\phi$ term from the operator $\Ho_{aa}^{(\alpha)}$:
\begin{equation}
\Ho_{aa}^{(\alpha)}[\phi]  -  \Delta\phi +2\left(1-\alpha\right)\nabla\phi\cdot\nabla U = -2 \rf\cdot\nabla\phi \,.
\end{equation}
Using \eqref{eq:h4^1} the above simplifies to 
\beq \label{eq:rphi}
(\Ho_{aa} ^{(\alpha)}  -  \Ho_{ss} ^{(\alpha)})[\phi] = -2 \rf\cdot\nabla\phi \,,
\eeq
which holds for every $\phi\in {\cal C}^2(\M)$ and for every $\alpha\in [0,1]$.
Our task is to intelligently choose a set of functions $\phi$ which plugged into \eqref{eq:rphi} will allow us to reconstruct $\rf$. In particular, it is easy to see that gradients of the chosenn $\phi$'s will have to span the tangent plane $T\M_x$ at any $x$.

 We  exploit the simple fact that the eigenvectors $\psi_k$ of $\Ho_{ss}^{(1)}$, with $k=2,\ldots d$ recover $\mathcal{M}$. This would mean they serve as coordinates, and as such $\nabla \psi_k = e_k$ on $\mathcal{M}$, that is the gradient of a coordinate function is the unit vector in the $k$ direction by definition. Hence, the $\psi_k$ used for $\phi$ in equation \eqref{eq:rphi}  recover the component of $\rf$ parallel to $e_k$. Hence, (using $\alpha = 1$), we have: 
\begin{equation}
(\Ho_{aa}^{(1)}  - \Ho_{ss}^{(1)}) [\psi_k] = -2 \rf\cdot e_k = -2 r_k \,. \label{eq:rk}
\end{equation}
In other words, it is straightforward to recover the component of $\rf$ that lie in tangent space $T\mathcal{M}$ at each point of $\mathcal{M}$ by applying $\Ho{aa}^{(1)}-H_{ss}^{(1)}$ to the coordinates obtained in step 6. This approach is implemented in steps 8--13 of Algorithm~\ref{alg:geo}. The matrix $P$ is the transition matrix of $A$, $T$ is the renormalized kernel, and $H_{aa}^{(1)}$ is the  discrete version of $\Ho_{aa}^{(1)}$ denoted by $\Ho_{aa,n}^{(\alpha=1)}$ in Figure \ref{fig:mod_sem}. 

The last step produces a $n\times d$ matrix $R$, where $R_{ij}$
represents component $j$ of the vector field $\rf$ (tangential) at
data point $i$.

The running time of the algorithm is of the same order of magnitude as
that of the Diffusion Map embedding, i.e.  ${\cal O}(n^3)$, as steps
1--7 essentially replicate this algorithm. The remaining steps contain
matrix renormalizations, adding a number of ${\cal O}(n^2)$ operations.

\begin{algorithm}[tb]
   \caption{Directed Embedding}
   \label{alg:geo}
\begin{algorithmic}
   \STATE {\bfseries Input:} Affinity matrix $A_{i,j}$ and embedding dimension $d$
   \STATE 1. $S \leftarrow (A+A^T)/2$
   \STATE 2. $q_i \leftarrow \sum_{j=1}^n S_{i,j}, \, \, Q = diag(q)$
   \STATE 3. $V \leftarrow Q^{-1}SQ^{-1}$
   \STATE 4. $q^{(1)}_i \leftarrow \sum_{j=1}^n V_{i,j}, \, \, Q^{(1)} = diag(q^{(1)})$
   \STATE 5. $H_{ss}^{(1)} \leftarrow Q^{(1)^{-1}}V$
   \STATE 6. Compute the the eigenvalues $2:d+1$  of $H_{ss}^{(1)}$ (in decreasing order of their value) and their right eigenvectors. Let $\Lambda$ by the $d \times d$ diagonal matrix of the eigenvalues, and $\Phi$ the $n\times d$ matrix formed by the eigenvectors. The columns of $\Phi$ are the $d$ coordinates of the embedding.
   \STATE 7. Compute $\pi$ the left eigenvector of $H_{ss}^{(1)}$ with eigenvalue 1.
   \STATE 8. $\pi \leftarrow \pi/\sum_{i=1}^n{\pi_i}$ is the density distribution over the embedding. 
   \STATE 9. $p_i \leftarrow \sum_{j=1}^n A_{i,j}, \, \, P = diag(p)$
   \STATE 10. $T \leftarrow P^{-1}A P^{-1}$
   \STATE 11. $p^{(1)}_i \leftarrow \sum_{j=1}^n T_{i,j}, \, \, P^{(1)} = diag(p^{(1)})$
   \STATE 12. $H_{aa}^{(1)} \leftarrow P^{(1)^{-1}}T$
   \STATE 13. $R \leftarrow (\Phi \Lambda - H_{aa}^{(1)}\Phi)/2$. The columns of $R$ are the vector field components.
\end{algorithmic}
\end{algorithm}

An alternate approach we considered is to estimate $c-2\nabla \cdot
\rf$. This is achieved by taking advantage of the fact that
\begin{equation} \label{eq:alternate-recovery}
(H_{as}^{(1)}-H_{ss}^{(1)})[\phi] = (c+2  \nabla \cdot r) \phi \,,
\end{equation}
is a diagonal operator, in the limit $\epsilon \to 0$. This approach suffers from the fact that $c$, the effect of the normal component of $\rf$, is unknown. So, it may work, estimating $\nabla \cdot \rf$, only if $\rf(x)$ has no component perpendicular to the tangent space $T\mathcal{M}$.

We have implemented this estimation, and we have also taken into
account that in practice $(H_{as}^{(1)}-H_{ss}^{(1)})[\phi]$ is not
perfectly diagonal. Hence we used $\phi = 1$ instead of simply
computing the diagonal of $(H_{as}^{(1)}-H_{ss}^{(1)})[\phi]$. Even
so,  recovering $\nabla
\cdot \rf$ was found to be numerically unstable as $\nabla \cdot r$ is very
sensitive to the spacing of point sampled on $\mathcal{M}$. This is
not too surprising given that discrete derivative are very sensitive
to grid spacing. Therefore, we are not pursuing this approach further.

\section{Experiments}

\subsection{Artificial Data}
\label{sec:experiments-artif}

\begin{figure*}
\vskip 0.2in
\begin{center}
\begin{tabular}{cc}
\includegraphics[width=0.45\textwidth]{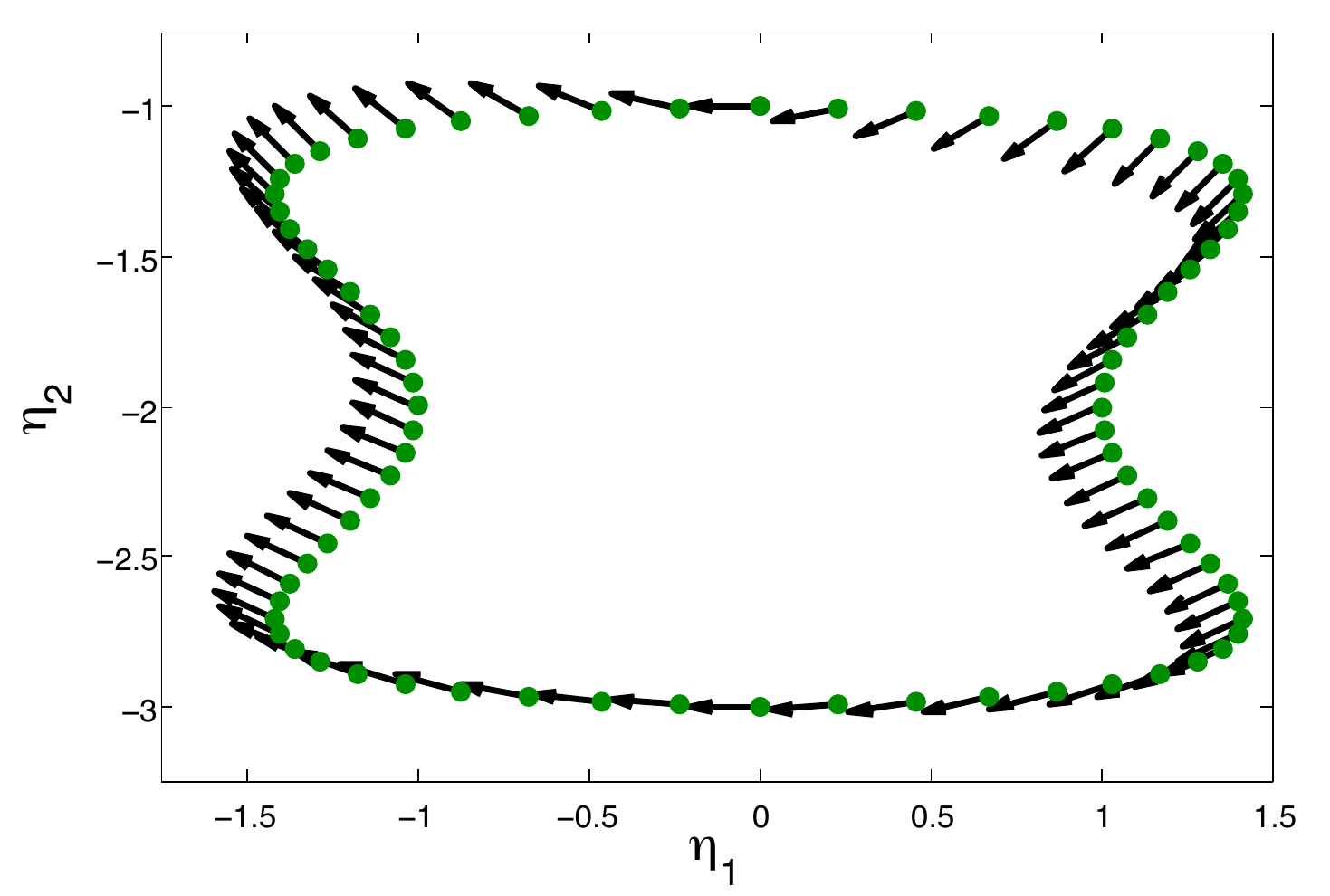} &
\includegraphics[width=0.45\textwidth]{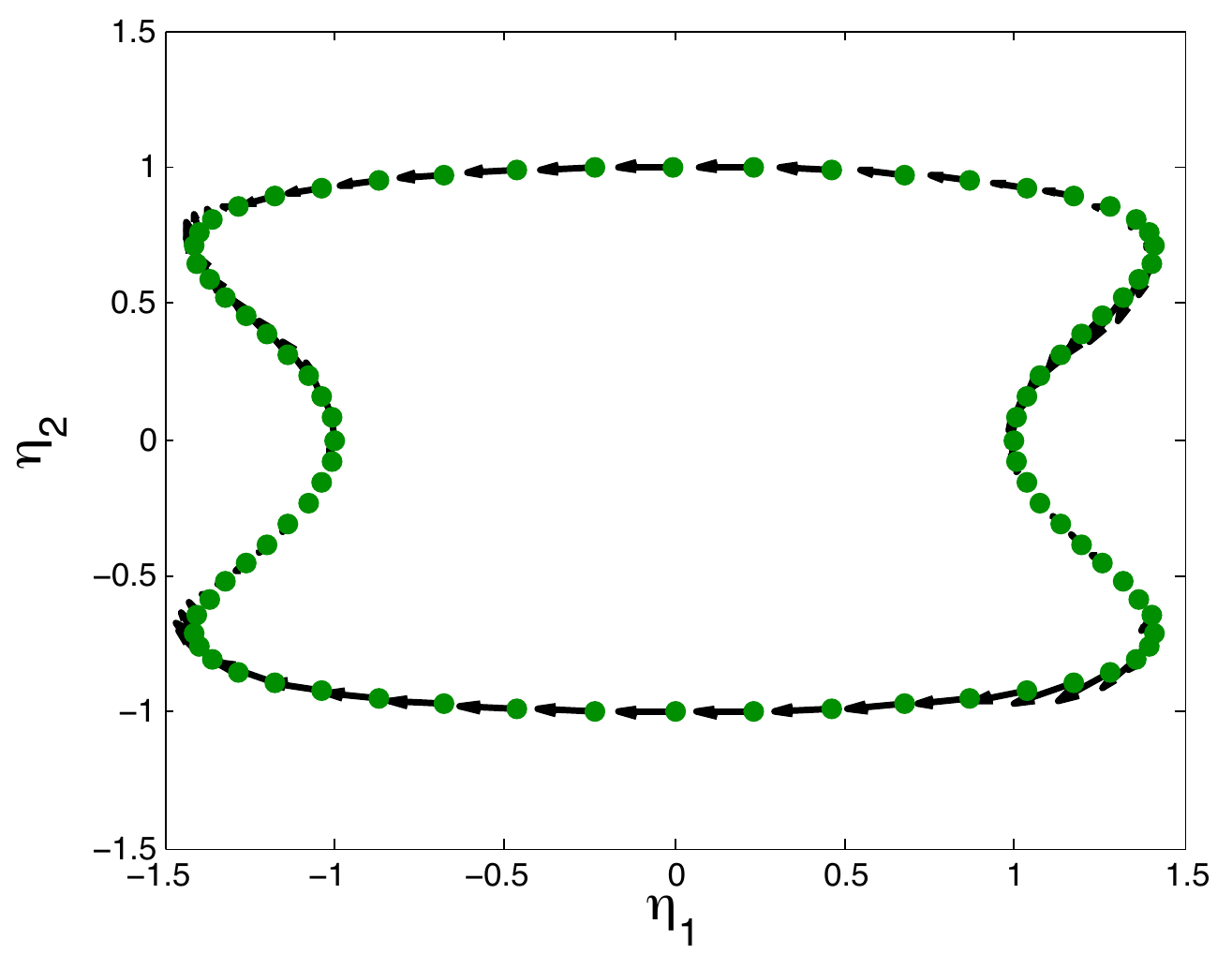} \\
\end{tabular}
\caption{Artificial data of a one dimensional manifold sampled on a uniform grid with a smoothly varying vector field (left) and its recovery (right). The vector field has a component normal to the manifold, which cannot be recovered. The algorithm correctly recovers  the tangential component of the vector field. we used $\epsilon = 0.03$ and 400 sample points.
}
\label{fig:art}
\end{center}
\vskip -0.2in
\end{figure*}

\begin{figure}
\setlength{\picwi}{0.23\textwidth}
\begin{tabular}{cccc}
\includegraphics[width=\picwi,bb=170 220 430 530,clip]{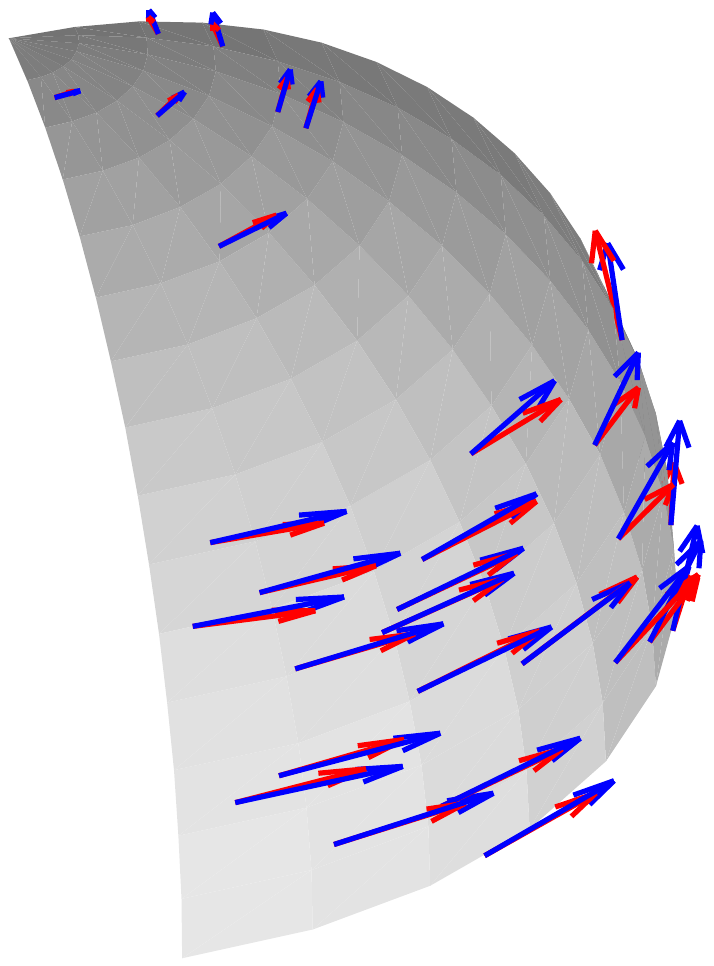}&
\includegraphics[width=\picwi,bb=170 220 430 530]{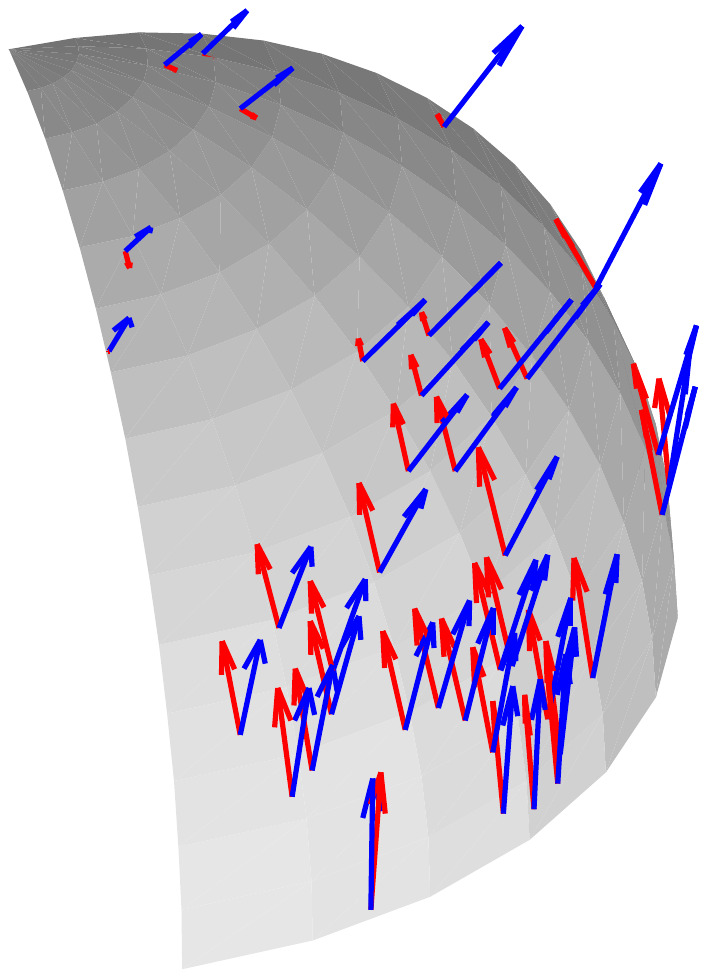}&
\includegraphics[width=\picwi,bb=170 220 430 530,clip]{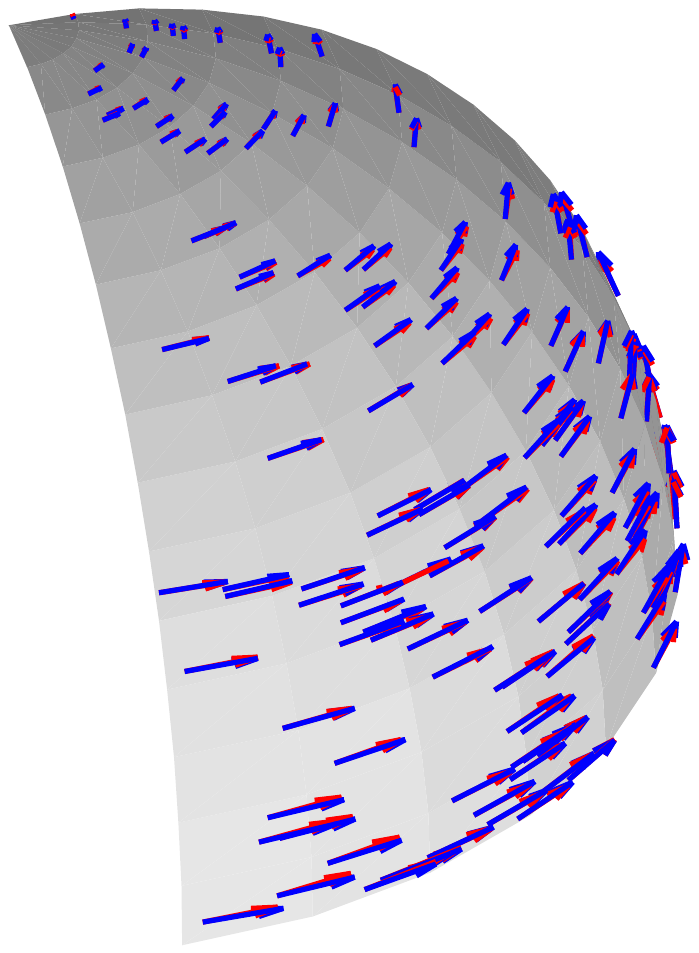}&
\includegraphics[width=\picwi,bb=170 220 430 530,clip]{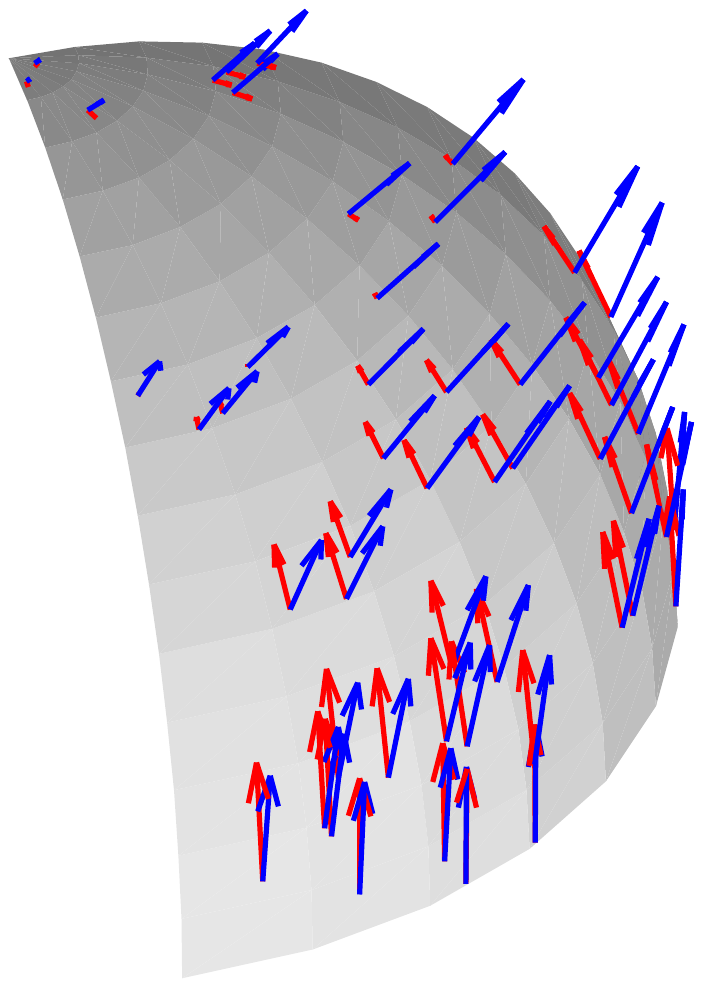}\\
\end{tabular}
\caption{Spherical octant with non-uniform sampling and vector field $\rf$ (in blue at 5\% of the sampling points). The recovered $R$ is shown in red. Tangential$\rf$ (left) and $\rf$ with normal component (right). Recovery from 500 samples (top), respectively 5000 (bottom).
}\label{fig:octant}
\end{figure}

For illustrative purposes, Figure~\ref{fig:art}, we begin by applying
our method to a simple example. We use one dimensional ``dumbbell''
manifold, with a regular sampling grid as recovering the density is
quite straight forward density distribution in the form of a sinus of
the circle. We exemplify a vector field that is not tangent to the
manifold to show how the algorithm recovers only the tangiential
component. The recovery of all three components is very good, even at
this low sample size. As expected, only the tangential component of
the vector field is recovered, as the normal component is
unidentifyable from the diffusion process alone. 
The next example is a 2 dimensional manifold. We experiment with 500
and 5000 sample points, and with two different vector fields, one of
them with a normal component and one restricted to the tangent plane
of $\M$. The sampling is non-uniform. Figure \ref{fig:octant} shows
only the vector field, for clarity. We see that even in the case of
small sample size the recovery of $\rf$ is remarkably accurate. 

The final illustrative experiment considers the more involved example
of a whole sphere with points sampled according to the Earth's
geographical map, with a vector field on it whose direction is tangent
to the latitude curves. The embedding algorithm maps now to three dimensions. 
\begin{figure*}
\includegraphics[bb=120 140 720 370,clip,height=2.5in,width=\textwidth]{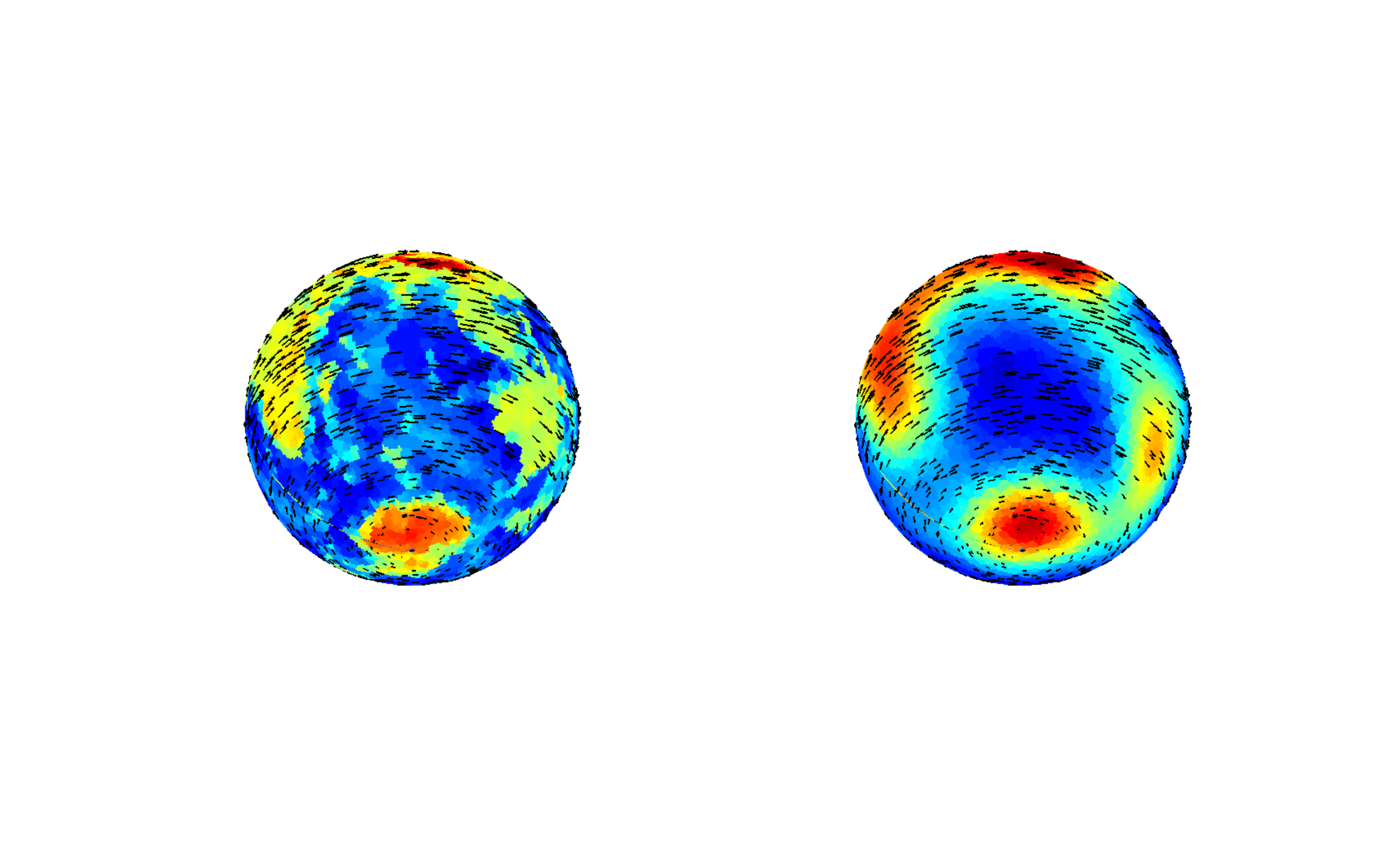}
\caption{Sphere with non-uniform map of sampling density and latitudinal vector field superimposed: model (left) and recovered components (right) from 4000 sample points with    $\epsilon = 0.01$.}\label{fig:tearth}
\end{figure*}

\subsection{Real Data}\label{sec:experiments-real}
 The National Longitudinal Survey of
Youth (NLSY) followed a representative sample of young men and women in the
United States from 1979 to 2000. This data was curated by \cite{}, producing the data set we will refer to as {\tt Jobs}.

The part of data set we use consists of a sample of 7,816 individual career
sequences, of length 64, listing the jobs a particular individual held
every quarter between the ages of 20 and 36. Each {\em token} in the
sequence identifies a job as an {\em industry $\times$ occupation}
pair. Approximately 213 distinct pairs occur in the data more than 50
times and we take these as our data set. They span 25 unique industry
and 20 unique occupation indices. Thus, our graph has 213 nodes - the
jobs - and our observations consist of 7,816 walks between the graph
nodes.

We convert these walks to a directed graph with weigh matrix $A$,
where $A_{ij}$ represents the number of times a transition from job
$i$ to job $j$ was observed and setting the diagonal to 0. This latter
step is motivated, in essence, by the observation that the ``sticky''
nature of jobs is an additional phenomenon, not modeled by our
framework, and in fact largely independent of job type. Practically,
we notice that small changes to the diagonal of $A$, e.g from 0 to 1,
do not affect the final result significantly. Moreover, the removed
diagonals can be preserved as a node attribute for further analysis.
Note that this matrix is {\em asymmetric}, i.e $A_{ij}\neq A_{ji}$.

\begin{figure*}
\vskip 0.2in
\begin{center}
\begin{tabular}{cc}
\includegraphics[width=0.45\textwidth,clip,bb=50 180 550 585]{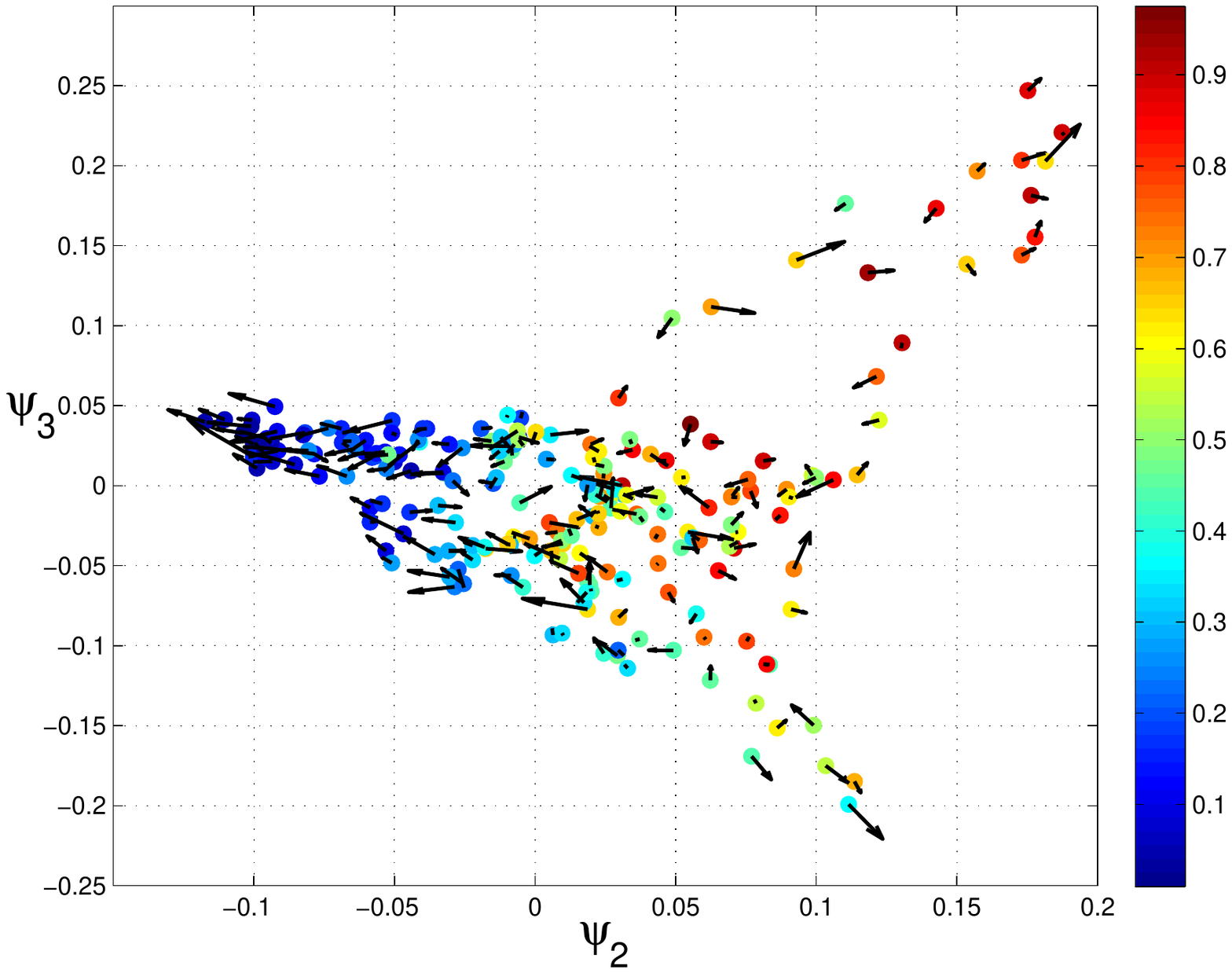}
&
\includegraphics[width=0.45\textwidth,clip,bb=50 180 550 585]{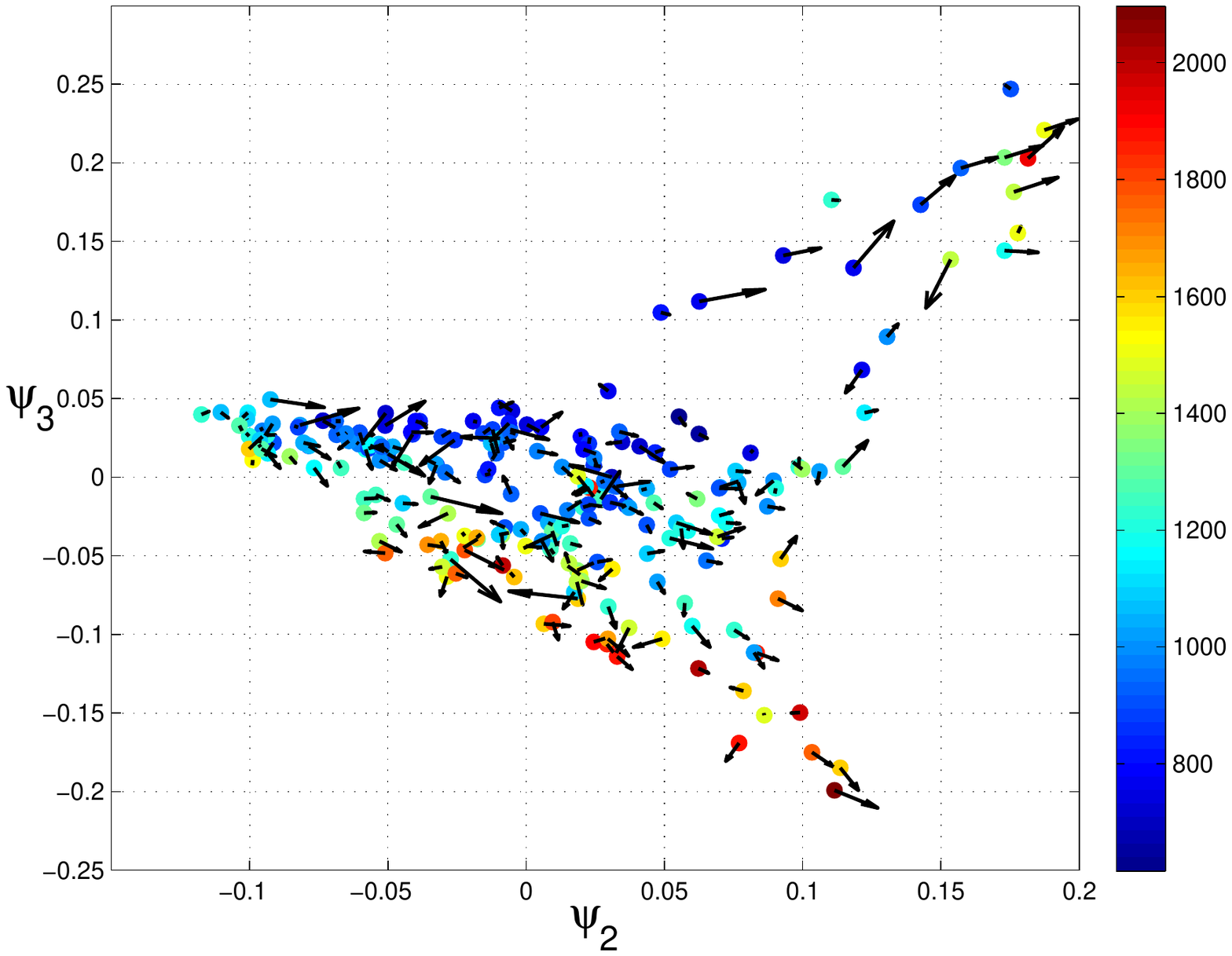}
\\
\end{tabular}
\caption{\label{fig:gender}
Embedding of {\tt Jobs} along with estimated vector fields: total flow $\nabla U+\rf$ (left) and $\rf$ alone (right). The color map represents gender proportion (left, male = 0 and female = 1) and monthly wages in dollars (right). 
}
\end{center}
\vskip -0.2in
\end{figure*} 

With  algorithm \ref{alg:geo} we obtain an embedding of the job market that describes the relative position between jobs, their distribution, and the natural time progression from each job. From a purely statistical perspective, the embedding maps the ``job'' variable, a discrete variable with a very high number of possible values, to a much more manageable continuous variable. Besides the pragmatic advantages of the continuous parametrization w.r.t the discrete one, this model is highly plausible, as it is conceivable that jobs form a continuum, in which every encoding is a form of quantization. Our embedding algorithm is thus recovering the relative positions of the jobs in this continuum. Along with this, the natural time progression in jobs space is highly  interesting.

 Together, they summarize the job market dynamics by describing which jobs are naturally close as well as where they can lead in the future. From a public policy perspective, this can potentially address important questions such as which job may require extra attention in terms of helping individuals attain better upward mobility. 

The job market was found to be a relatively high dimensional manifold,
with about 10--12 eigevectors with significant eigenvalue. We present
here only the dimensions that showed correlation with important
demographic data, such as wages and gender. These were found to be
coordinates 2 and 3, while other coordinates did not have a clear
meaning. Figure~\ref{fig:gender} shows this two-dimensional
sub-embedding along with the directional information for each
dimension. We see the evident variation of gender along $\Psi_2$ and
of wages along $\Psi_3$.  The plots show a strong outward progression
for most jobs, where outward progression is correlated to increasing
wages.

To note that for describing the job dynamics, we are interested in the
total advective flow, that sums the contributions of $\rf$ and of
diffusion, i.e $2(\nabla U +\rf)\cdot\nabla\phi$. It is easy to see
that this flow can be obtained from $\Ho_{aa}^{(0)}-\Ho_{ss}^{(1)}$,
or, in the matrix notation of Algorithm
\ref{alg:geo}, as $(\Phi\Lambda-P\Phi)/2$. This is the vector field
plotted in the left of Figure \ref{fig:gender}, while $\rf$ alone is
plotted on the right.

The temporal progression shown is explained by the natural diffusion process due to the non-uniform sampling density in job space, augmented with  $\rf$ which represents the other economic forces at work. From a policy-maker's perspective, it is the latter term that matters, and that could be used for intervention, therefore $\rf$ is depicted in Figure \ref{fig:gender}.

The embedding (left panel) suggests that males (blue) and females
(red) tend to move away from each other over time, especially in the
region of low-paying jobs, which seem to be specialized by
gender. There is a particularly strong observed flow, mostly explain
by the diffusion term, for men in low paying jobs to segregate towards
more gender segregated, but slightly better paying jobs (e.g
construction worker). One notices too a flow towards the lower corner,
characterized by high paying jobs with equal proportions of men and
women. The upper right corner of the embedding is occupied by women
specific jobs. A few of these are high-paying (nurse). In this corner,
a very strong $\rf$ (flow explained by economic forces different than
diffusion) leads from very low paying jobs towards the higher end
jobs. It is to investigate if this shows that many women in low paying
jobs, unlike the men in the same situation, are subject to some form
of outside force towards upward mobility, but in reality (total flow)
benefit little from it.

One notes also that the low-wage jobs away from the corners (in blue
on the right panel) show little overall progression, while higher paid
jobs seem to show some progression towards even better wages.

\section{Discussion} 
\label{sec:discussion}

From the theoretical point of view, we have introduced a generative
model for directed graphs, in which the asymmetry is modeled
explicitly as a vector field $\rf$; this is a plausible and general
model for digraphs. Moreover, we have shown that our representation
for asymmetry is almost optimal, in the sense that representation that
is more general is not identifiable from the data. This model is the
first attempt to represent directed affinities with an infinitezimal
generative model on a manifold\footnote{By this, we mean a generative
model where the affinities are discrete version of an infinitezimal
transport $k(x,y)$. The literature is rich in other probabilistic
models of random graphs, of a combinatorial nature.}

Second, with Theorem \ref{thm:lim} we proved the continuous limit of a
large class of operators based on asymmetric kernels. This is an
original contribution onto itself.

As an example, our main theorem allowed us to  derive the
continuous limit of the WCut algorithm. The relationship between the
WCut and this work parallels that between the MNCut of
\cite{ShiMalik_ncut_pami:00}, and \cite{dmap06},~\cite{belniy02}. The latter describe a generative process and operators that can be thought as the continous limits of the directed graph, random walk, graph Laplacian, and embedding
algorithm. This paper performs the same analysis for the WCut
algorithm as an embedding method for digraphs. Note too that that the
WCut we use is similar to \cite{zhou:05b}. One could perform similar
limit analyses based on Theorem \ref{thm:lim} for other existing or
novel directed embedding algorithms.

The theoretical results form the basis for a spectral algorithm that
can be applied to a directed graph, to produce an embedding in a
low-dimensional manifold, an estimate of the data density on the same
manifold, and an estimate of the local directional flow at each point,
which is the source of the graph's asymmetry. As a byproduct of
separating the asymmetry of the graph in the last feature, the
directional flow, the geometry of the resulting manifold is identical
to the one obtained by embedding a symmetric graph using the Diffusion
Map algorithm of \cite{dmap06}. We underscore that this property of
our algorithm is not an ad-hoc choice, but a direct consequence of
using a principled approach based on a generative model. Since the
generative model for the embedding is the model proposed by
\cite{dmap06}, it is only natural that our algorithm recovers the same coordinates and sampling distribution as the Diffusion Map algorithm. 

We have used from \cite{dmap06} the idea to renormalize the kernel (by
$\alpha$) in order to be able to separate the density $p$ and the
manifold geometry $\mathcal{M}$, and used some of the technical
results developed there in our proofs. However, with respect to the
{\em asymmetric part} of the kernel, everything starting from the
defintion and the introduction of the vector field $\rf$ as a way to
model the asymmetry, through the derivation of the asymptotic
expression for the symmetric and asymmetric kernel, is new. Regarding
the use of renormalization, we go significantly beyond the elegant
idea of \cite{dmap06} by analyzing the four different renormalizations
possible for a given $\alpha$, each of them combining different
aspects of $\mathcal{M},p$ and $\rf$. Only the successful combination of
two different renormalizations is able to recover the directional flow
$\rf$.

With respect to the previous work on directed graph embedding, several
algorithms have been proposed for the embedding of directed graphs in
low dimensional manifolds \cite{PentneyM:aaai05,penmei07,zhou:05b,
chung:conf}. However, the analysis that relates a graph algorithm with
a continuous generative model has not so far been attempted for
directed graphs. In particular, the above methods do not explicitly
attempt to separate the source of the asymmetry from the
embedding. Hence, they only output a graph embedding, with no analog
of the $\rf$ estimate to explicitly account for the observed
asymmetry. Therefore, just like in the embedding of undirected graphs,
a non-uniform sampling density, if not accouted for explicitly, will
confound the embedding coordinates, the unaccounted for asymmetry will
appear in the embedding of the graph.
\comment{show this in figure}

\cite{ting10} (THJ) presented asymptotic results for a general
family of kernels that includes asymmetric and random kernels. Our
$k_\epsilon$ can be expressed in the THJ notation by taking
$w_x(y)\leftarrow 1+\rf(x,y)\cdot(y-x)$, $r_x(y)\leftarrow 1$,
$K_0\leftarrow h$, $h\leftarrow \epsilon$. The assumptions in TJH are
more general than the assumptions we make here. However, THJ focus on
the graph construction methods (seen as generative models for graphs),
while we focus on explaining {\em observed directed graphs} through a
manifold generative process. Moreover, while the THJ results {\em can}
be used to analyze data from directed graphs, they differ from our
Proposition \ref{thm:lim}. Specifically, with respect to the limit
in Theorem 3 from THJ, we obtain the additional terms $f(x)\nabla\cdot
\rf$ and $c(x)f(x)$ in Theorem \ref{thm:lim}.

How general is our setup? The view of the graph nodes as sampled from
a continuous manifold is standard in the machine learning spectral
graph theory. However, for directed graphs, to the best of our
knowledge, no one has before attempted to extend this generative
model. In this respect the reader should note that our model of
asymmetry is the most general possible, assuming the unidentifyable
components of the $\rf$ are set to fixed values. These are the
component of $\rf$ perpendicular to the manifold, whose effect is the
function $c$ in Theorem \ref{thm:lim}, and a multiplicative constant
for $\rf$, resulting from the fact that the coordinate functions
$\Phi$, as eigenvectors of the renormalized Laplacian $H^{(1)}_{ss}$,
can only be recovered up to a multiplicative constant.

\comment{Regarding the algorithmic novelty, the reconstruction the manifold
geometry relies on existing algorithms. The new contribution
includes the well-founded algorithm for recovering the vector field,
but it also includes combining what amounts to existing
methods that allow us to separate the vector field from the other
model components and to represent it correctly. In the supplementary materials, we demonstrate that these methods/choices cannot be arbitrary. }

We note here that recovering $\rf$ is just one, albeit the most
useful, of the many ways of exploiting these theoretical resuls. For
instance, one can also recover $\nabla\cdot \rf$ directly, obtain the
WCut embedding for $\alpha=0$, and check the consistency of the model
on the real example at hand.

We foresee several other immediate applications of the framework
developed here, like checking the model fit, estimating the dimension,
parametrizations of the components $\mathcal{M},p,\rf$ and so on.  The
practical applications are immediate, and we are currently applying
this method to the analysis of the career sequences of
section~\ref{sec:experiments-real}. Since the different coordinates of
the manifold correlate well with demographic parameters like sex,
wages, education, we expect that the different components of the
vector field to be interpretable as social and personal forces that
drive the individual carrer paths at different points in the job
space.

\appendix
\section*{Proof of the operator expansion \eqref{eq:asaim}\label{app:asymptotic}}

The core of the proof is the asymptotic expansion of both $k_\epsilon$
and $f$ around a point $x\in \M$. In the same time we exploit the
 decay of the kernel $h_\epsilon$ with distance to
approximate the integral over $\M$ with an integral over a ball around
$x$ with radius of order $\sqrt{\epsilon}$.  This allows for
integrating the kernel over the tangent space $T\mathcal{M}_{x}$ at a
given point $x\in\mathcal{M}$. Thus, the integration
in~\eqref{eq:asaim} can be now changed from $\mathcal{M}$ to $\RR^{d}$
because on a ball around $x$ where $\mathcal{M}$ can be approximated
to $T\mathcal{M}_{x}$ (which is isomorphic to $\RR^{d}$).
\begin{eqnarray} \label{eq:dasym}
\lefteqn{\int_{\mathcal{M}} k_{\epsilon}(x,y)f(y)dy
 \;=\; \int_{\RR^{d}}k_{\epsilon}(x,x+(u,g(u)))\times}\\
&&\left(\bar{f}(0)+\sum_{i=1}^{d}u_{i}\frac{\partial\bar{f}(0)}{\partial s_{i}} 
 +\frac{1}{2}\sum_{i,j=1}^{d}u_{i}u_{j}\frac{\partial^{2}\bar{f}(0)}{\partial s_{i}\partial s_{j}}
  +\mathcal{Q}_{x,3}(u)\right)
\nonumber\\
&&\times\left(1+\mathcal{Q}_{x,2}(u)
 +\mathcal{Q}_{x,3}(u)\right)du+O(\epsilon^{2})\,.\nonumber \end{eqnarray}
In the derivation above, we used a number of asymptotic results
of~\cite{dmap06}. We do not reproduce these here at length, as they
can be found in Appendix B of~\cite{dmap06}.  We now explain the
origin of each of the terms in the expansion. The main idea is finding
a natural set of coordinates to exploit the fact that the manifold
$\mathcal{M}$ is locally flat.

The $s_{i}$'s are normal coordinates along geodesics of $\mathcal{M}$
that are defined by an orthonormal basis $(e_{1},...,e_{d})$ of the tangent space $T\mathcal{M}_{x}$.
These coordinates serve to establish that the derivatives are taken
along orthogonal geodesics of $\mathcal{M}$ at $x$, and so they can
be used to define differential operators on $\mathcal{M}$ such as
the Laplacian: \begin{equation}
\Delta f(x)=\sum_{i=1}^{d}\frac{\partial^{2}\bar{f}(0)}{\partial s_{i}^{2}}\,. \nonumber \label{eq:laplacian}
\end{equation}
 Meanwhile, the $u_{i}'s$ are the coordinates of $T\mathcal{M}_{x}$
in the $(e_{1},...,e_{d})$ basis. That is, they are determined by
the projection of the manifold onto the $T\mathcal{M}_{x}$ by $u_{i}=(y-x)\cdot e_{i}$.
The manifold is then locally parameterized by $y=x+(u,g(u))$ with
$g:\RR^{d}\rightarrow\RR^{n-d}$ with $g(0)=0$ and $\frac{\partial g(0)}{\partial u_{i}}=0$
by construction. Finally, $\mathcal{Q}_{x,n}(u)$ denotes a polynomial in $u$ of order $n$ at $x$.

It remains to explain how two of the terms in~\eqref{eq:dasym} arise. The derivatives of $\bar{f}$ result from a Taylor expansion
of $f(y)$ along the geodesics parametrized by the coordinates $(s_{1},...,s_{d})$.
The Taylor expansion is then expressed in terms of the projected coordinates $u_{i}$ by making use of the fact that $s_{i}=u_{i}+\mathcal{Q}_{x,3}(u)+O(\epsilon^{2})$ (as the kernel radius is of the order $\sqrt{\epsilon}$ to bound the error by $\epsilon^2$ we need to go up to a third degree polynomial in the expansion). Meanwhile, the polynomial $1+\mathcal{Q}_{x,2}(u)+\mathcal{Q}_{x,3}(u)$ before the differential $du$ comes from the volume element of $\mathcal{M}$
at $y=x+(u,g(u))$. Specifically, it originates from the change of
variable $y\rightarrow u$ given by $|det(dy/du)|=1+\mathcal{Q}_{x,2}(u)+\mathcal{Q}_{x,3}(u)$.

Next, we expand $k_{\epsilon}(x,x+(u,g(u)))$ in \eqref{eq:dasym}
in $u$. Dropping the terms that will be of order higher than $\epsilon$, we obtain: 
\begin{eqnarray}
h(||y-x||^{2}/\epsilon =h(||u||^{2}/\epsilon)
+\frac{\mathcal{Q}_{x,4}(u)}{\epsilon}h'(||u||^{2}/\epsilon)
+O(\epsilon^{3/2})\,, \label{eq:hasym}
\end{eqnarray}
 and 
\begin{eqnarray}\label{eq:aasym}
\lefteqn{\rf(x,y)\cdot(y-x) \,=\,}\\
&& (\bar{r}(x,0)+\sum_{i=1}^{d}u_{i}\frac{\partial\bar{r}(x,0)}{\partial s_{i}})
 \cdot(u,\sum_{i,j=1}^{d}\frac{u_{i}u_{j}}{2}\frac{\partial^{2}g(0)}{\partial u_{i}\partial u_{j}})+O(\epsilon^{2})\,.\nonumber \end{eqnarray}
As mentioned before, $\rf(x,y)$ is taken here to represent a vector
field on $\mathcal{M}$. To make sense of this interpretation, we
note that the $y$ variable is important only in a ball of radius
$\epsilon^{1/2}$ around $x$ because of the form of $h_{\epsilon}(x,y)$,
and that only the linear term in the expansion of $\rf(x,y)$ is important.
If we impose that $\nabla_{x}\rf(x,y)|_{y=x}=\nabla_{y}\rf(x,y)|_{y=x}$,
it follows that the kernel ``sees'' locally in the same vector
field as the global vector field $\rf(x,x)$ defined on $\mathcal{M}$.

Substituting~\eqref{eq:hasym} and~\eqref{eq:aasym} into~\eqref{eq:dasym}
and integrating finishes the proof. 

The first two terms in~\eqref{eq:w}, $\omega(x)f(x)$ and $\Delta
f(x)$, come from the symmetric kernel $h_{\epsilon}(x,y)$. $\Delta
f(x)$ comes from
\begin{equation}
\sum_{i,j=1}^{d}\int_{\RR^{d}}u_{i}u_{j}\frac{\partial^{2}\bar{f}(0)}{\partial s_{i}\partial s_{j}}\frac{h(||u||^{2}/\epsilon)}{\epsilon^{d/2}}du\,=\,\epsilon\Delta f(x)\,,\nonumber
\end{equation}
where all the off-diagonal terms in the sum vanish because the kernel was chosen to be radially symmetric. Meanwhile, $\omega(x)f(x)$ is
a geometric effect of the manifold $\mathcal{M}$, as it embodies the
effect of the manifold on the kernel and the change in volume element. As this term ultimately cancels out, it will not be further discussed.

The remaining terms are all contributions from the asymmetric part
of the kernel. The $\rf\cdot\nabla f(x)$ term comes from the kernel
picking up transport induced by the vector field $\rf(x)=\rf(x,x)$ \begin{equation}
\epsilon \rf\cdot\nabla f(x)=\sum_{i,j=1}^{d}r_i\int_{\RR^{2}}u_{i}u_{j}\frac{\partial\bar{f}(0)}{\partial s_{j}}\frac{h(||u||^{2}/\epsilon)}{\epsilon^{d/2}}du\,, \nonumber
\end{equation}
 while $f(x)\nabla\cdot \rf$ comes from the kernel augmenting or reducing
$f(x)$ by the amount of divergence of the vector field 
\begin{eqnarray}
\lefteqn{\epsilon f(x)\nabla_{z}\cdot \rf(z,x)|_{z=x}\,=\,}\nonumber \\
&&\sum_{i,j=1}^{d}\int_{\RR^{2}}u_{i}u_{j}\frac{\partial\bar{r_{j}}(x,s)}{\partial s_{i}}|_{s=0} 
 \times \frac{h(||u||^{2}/\epsilon)}{\epsilon^{d/2}}du\,, \nonumber
\end{eqnarray}
 where we have implicitly made use of the fact that $r$ is assumed
to satisfy $\nabla_{x}\rf(x,y)|_{y=x}=\nabla_{y}\rf(x,y)|_{y=x}$. The
remaining term, $c(x)f(x)$, arises if the vector field $\rf(x)$ has
a component perpendicular to $T\mathcal{M}_{x}$ at a point where the orientation of $T\mathcal{M}_{x}$ is changing: 
\begin{equation}
\epsilon f(x)c(x)=\sum_{i,j,k=1}^{d}r_k\int_{\RR^{2}}u_{i}u_{j}\frac{\partial^{2}g_{k}(0)}{\partial u_{i}\partial u_{j}}\frac{h(||u||^{2}/\epsilon)}{\epsilon^{d/2}}du\,. \nonumber \end{equation}

\comment{
\begin{equation}
\lim_{\epsilon\to 0}\int_{\M}k_{\epsilon}(x,y)f(y)dy=m_{0}f(x)\,,\end{equation}
as required to ensure~\eqref{eq:infop} will lead to the correct asymptotic in the asymmetric case (with $m_0=1$ w.l.o.g).
}

\bibliography{arxiv14-directed-embedding}

\begin{thebibliography}{14}
\providecommand{\natexlab}[1]{#1}
\providecommand{\url}[1]{\texttt{#1}}
\expandafter\ifx\csname urlstyle\endcsname\relax
  \providecommand{\doi}[1]{doi: #1}\else
  \providecommand{\doi}{doi: \begingroup \urlstyle{rm}\Url}\fi

\bibitem[Andersen et~al.(2007)Andersen, Chung, and Lang]{chung:conf}
Reid Andersen, Fan R.~K. Chung, and Kevin~J. Lang.
\newblock Local partitioning for directed graphs using pagerank.
\newblock In \emph{WAW}, pages 166--178, 2007.

\bibitem[Belkin and Niyogi(2002)]{belniy02}
Belkin and Niyogi.
\newblock Laplacian eigenmaps for dimensionality reduction and data
  representation.
\newblock \emph{Neural Computation}, 15:\penalty0 1373--1396, 2002.

\bibitem[Coifman and Lafon(2006)]{dmap06}
Coifman and Lafon.
\newblock Diffusion maps.
\newblock \emph{Applied and Computational Harmonic Analysis}, 21:\penalty0
  6--30, 2006.

\bibitem[Coifman et~al.(2005)Coifman, Lafon, Lee, Maggioni, Warner, and
  Zucker]{coif05}
Coifman, Lafon, Lee, Maggioni, Warner, and Zucker.
\newblock Geometric diffusions as a tool for harmonic analysis and structure
  definition of data: Diffusion maps.
\newblock In \emph{Proceedings of the National Academy of Sciences}, pages
  7426--7431, 2005.

\bibitem[Hoff et~al.(2002)Hoff, Raftery, and Handcock]{hoffhanraf02}
Hoff, Raftery, and Handcock.
\newblock Latent space approaches to social network analysis.
\newblock \emph{Journal of the American Statistical Association}, 97:\penalty0
  1090--1098, 2002.

\bibitem[Meila and Pentney(2007)]{penmei07}
Meila and Pentney.
\newblock Clustering by weighted cuts in directed graphs.
\newblock In \emph{SIAM Data Mining Conference}, 2007.

\bibitem[Meil\u{a} and Shi(2001)]{MShi:aistats01}
Marina Meil\u{a} and Jianbo Shi.
\newblock A random walks view of spectral segmentation.
\newblock In T.~Jaakkola and T.~Richardson, editors, \emph{Artificial
  Intelligence and Statistics AISTATS}, 2001.

\bibitem[Nadler et~al.(2006{\natexlab{a}})Nadler, Lafon, and Coifman]{fp05}
Nadler, Lafon, and Coifman.
\newblock Diffusion maps, spectral clustering and eigenfunctions of
  fokker-planck operators.
\newblock In \emph{Neural Information Processing Systems Conference},
  2006{\natexlab{a}}.

\bibitem[Nadler et~al.(2006{\natexlab{b}})Nadler, Lafon, Coifman, and
  Kevrekidis]{dsys06}
Nadler, Lafon, Coifman, and Kevrekidis.
\newblock Diffusion maps, spectral clustering and reaction coordiantes of
  dynamical systems.
\newblock \emph{Applied and Computational Harmonic Analysis}, 21:\penalty0
  113--127, 2006{\natexlab{b}}.

\bibitem[Pentney and Meil\u{a}(2005)]{PentneyM:aaai05}
William Pentney and Marina Meil\u{a}.
\newblock Spectral clustering of biological sequence data.
\newblock In Manuela Veloso and Subbarao Kambhampati, editors,
  \emph{Proceedings of Twentieth National Conference on Artificial Intelligence
  (AAAI-05)}, pages 845--850, Menlo Park, California, 2005. The AAAI Press.

\bibitem[Shi and Malik(2000)]{ShiMalik_ncut_pami:00}
Jianbo Shi and Jitendra Malik.
\newblock Normalized cuts and image segmentation.
\newblock \emph{PAMI}, 2000.

\bibitem[Ting et~al.(2010)Ting, Huang, and Jordan]{ting10}
Ting, Huang, and Jordan.
\newblock An analysis of the convergence of graph {Laplacians}.
\newblock In \emph{International Conference on Machine Learning}, 2010.

\bibitem[Zhou et~al.(2005{\natexlab{a}})Zhou, Huang, and Scholkopf]{zhou:05a}
Zhou, Huang, and Scholkopf.
\newblock Learning from labeled and unlabeled data on a directed graph.
\newblock In \emph{International Conference on Machine Learning}, pages
  1041--1048, 2005{\natexlab{a}}.

\bibitem[Zhou et~al.(2005{\natexlab{b}})Zhou, Schoelkopf, and
  Hofmann]{zhou:05b}
Zhou, Schoelkopf, and Hofmann.
\newblock Semi-supervised learning on directed graphs.
\newblock In \emph{Advances in Neural Information Processing Systems},
  volume~17, pages 1633--1640, 2005{\natexlab{b}}.

\end{thebibliography}

\end{document}